\documentclass{article} 
\usepackage{iclr2027_conference,times}

\usepackage{amsmath,amsfonts,bm}

\def\eqref#1{equation~\ref{#1}}

\def\1{\bm{1}}

\DeclareMathAlphabet{\mathsfit}{\encodingdefault}{\sfdefault}{m}{sl}
\SetMathAlphabet{\mathsfit}{bold}{\encodingdefault}{\sfdefault}{bx}{n}

\usepackage{amsmath}
\usepackage{amssymb}
\usepackage{hyperref}
\usepackage{graphicx}
\usepackage{adjustbox}
\usepackage{booktabs}
\usepackage{capt-of}
\usepackage{url}
\renewcommand{\thefootnote}{\fnsymbol{footnote}}

\title{Omni-RS: Unifying Multi-Task Understanding and Generation in Remote Sensing}

\author{
\normalfont
\begin{tabular}{@{}l@{}}
\textbf{Pan Yin$^{*}$,\quad
Shuai Yang$^{*}$,\quad
Yuanhao Liu,\quad
Datao Tang,\quad
Jialin Li,\quad
Hao Wang,}\\
\textbf{Pei Liu,\quad
Kaiyu Li,\quad
Cong Zhao,\quad
Xiangyong Cao$^{\dagger}$}
\\[3pt]
Xi'an Jiaotong University \\[1pt]
\texttt{caoxiangyong@mail.xjtu.edu.cn}
\end{tabular}
}

\iclrfinalcopy 
\begin{document}
\pagestyle{empty}
\maketitle
\thispagestyle{empty}


\begingroup
\renewcommand{\thefootnote}{}
\footnotetext{\hspace{-1.5em}$^{*}$These authors contributed equally.}
\footnotetext{\hspace{-1.5em}$^{\dagger}$Corresponding author.}
\endgroup


\begin{abstract}

Practical remote sensing applications often require both understanding and generation capabilities for the same geographic scene. However, most existing methods address either understanding or generation in isolation, while existing unified remote sensing models' generative ability is limited to text-to-image. To bridge this gap, we propose Omni-RS, a unified framework supporting four remote sensing understanding tasks and five generation tasks. Specifically, we introduce the ``Scene-Guided Two-Stage Training'' strategy, which first adapts a multimodal large language model to remote sensing understanding and then uses its hidden states to guide a diffusion transformer for image generation. To enable joint learning across diverse generation tasks, we construct UniGen, a unified multi-task generation dataset, by organizing heterogeneous remote sensing data into a unified format. Furthermore, we introduce ``Detailed Visual Conditioning'', which uses visual conditions encoded by a VAE to guide the generation process, thereby supporting additional generation tasks that require reference images. Extensive experiments demonstrate that Omni-RS performs exceptionally well in understanding tasks and has achieved outstanding results in a variety of generative tasks.

\end{abstract}

\section{Introduction}

Remote sensing (RS) imagery serves as a foundational pillar for Earth observation, powering critical downstream applications such as disaster response, environmental monitoring, and urban planning~\citep{van2013remote,li2020review,wellmann2020remote}. Powered by recent advances in deep learning, the field has witnessed remarkable progress across two distinct fronts: visual understanding has evolved from basic classification to complex visual question answering, while generative modeling has expanded from simple text-to-image synthesis to practical tasks like inpainting and cross-modal translation~\citep{maxwell2018implementation,lobry2020rsvqa,chen2026rsedit}. However, these abilities have largely developed in their own isolated domains. In reality, real-world geospatial applications often require joint understanding and generation for a single scene, such as assessing damaged structures and inpainting missing imagery during disaster response. Consequently, developing a unified architecture that harmonizes multi-task RS understanding and generation is essential to addressing these intrinsically complementary needs.



\begin{figure*}[h]
  \centering
   \includegraphics[width=1\linewidth]{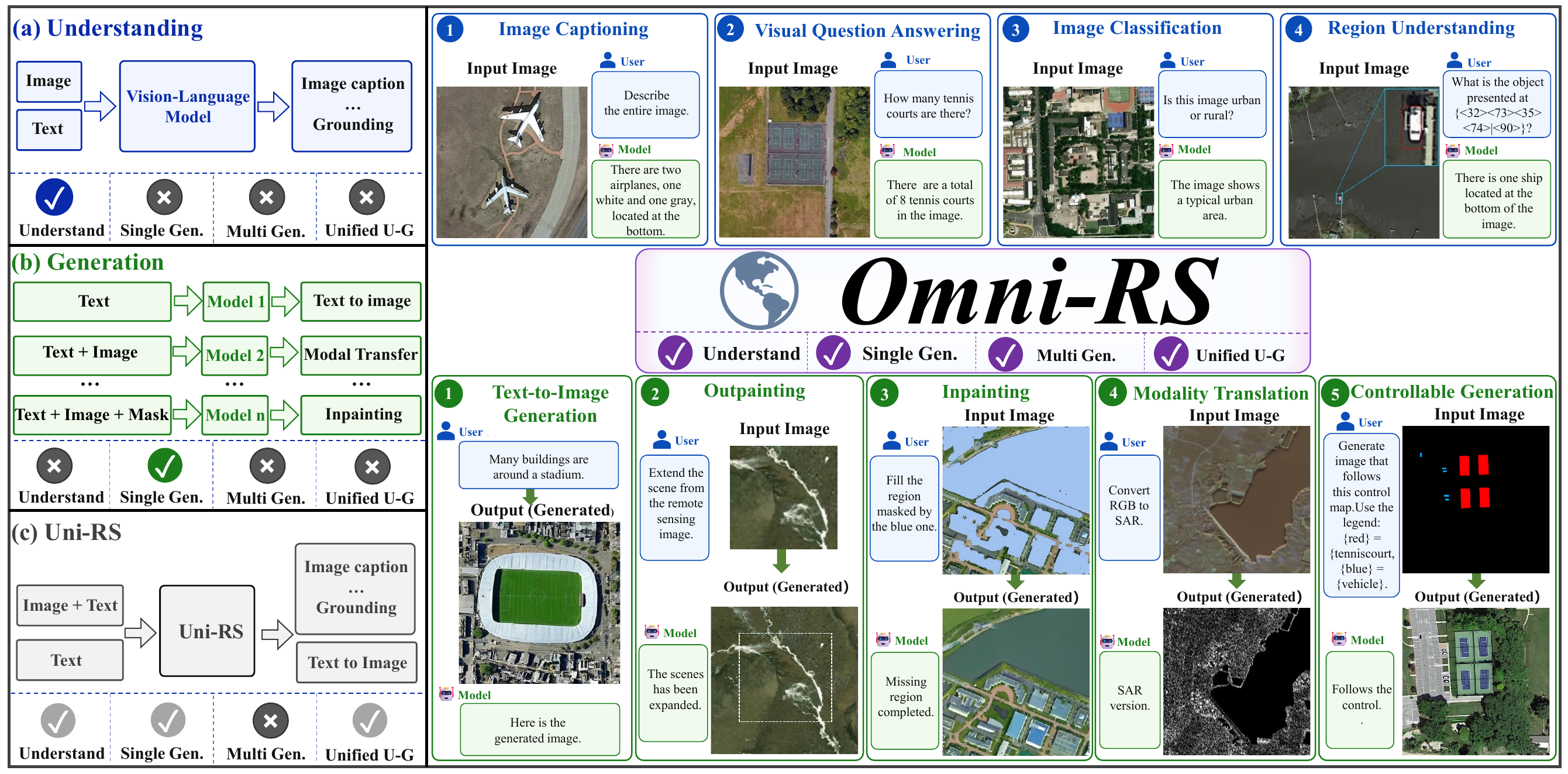}
   \caption{\textbf{Comparison of task capabilities across different remote sensing model paradigms.} (a) Understanding-based methods lack image generation capabilities. (b) Generation-based methods support individual generation tasks but lack understanding
and multi-task generation. (c) Uni-RS bridges understanding and generation, but remains restricted to text-to-image synthesis. Right: Omni-RS supports four understanding tasks and five generation tasks within a unified framework.}
   \label{fig:intro}
  \vspace{-10pt}
\end{figure*}

As illustrated on the left of Fig.~\ref{fig:intro}, existing remote sensing models generally fall into three categories: understanding-based, generation-based, and unified understanding-generation approaches. Understanding-based methods, such as GeoChat~\citep{kuckreja2024geochat} and SkySenseGPT~\citep{luo2024skysensegpt}, leverage instruction tuning or scene graph reasoning to enable region-level grounding and spatial analysis. While effective for visual comprehension, these models lack generative capabilities. Conversely, generation-based methods focus on image synthesis; for instance, CRS-Diff~\citep{tang2024crs} integrates multi-modal conditions for controllable synthesis, and RSEdit~\citep{chen2026rsedit} adapts diffusion models for image editing. Nevertheless, these generative architectures cannot perform understanding tasks. Additionally, varying conditioning inputs across different generation tasks hinder joint training, leaving most methods tailored to a single task. Recently, Uni-RS~\citep{zhang2026uni} connected a multimodal large language model (MLLM) to a diffusion generator via learnable queries, achieving text-to-image synthesis alongside visual understanding. However, its generative capability is still restricted to text-to-image synthesis, failing to support tasks that require additional reference images, such as inpainting and layout controllable generation.

Despite these advances, unifying diverse remote sensing understanding and generation tasks within a single model still presents three key challenges. First, understanding and generation impose fundamentally different optimization requirements. Although both tasks rely on shared scene semantics, understanding aims to interpret visual content into sparse textual tokens, whereas generation focuses on synthesizing spatially consistent visual pixels. Jointly optimizing these heterogeneous objectives often leads to task interference during training, hindering the model from mastering both capabilities simultaneously. Second, existing generation datasets remain highly fragmented. For example, Git-10M~\citep{liu2025text2earth} provides image-text pairs tailored for text-to-image synthesis, while TerraGen~\citep{tang2026terragen} couples images with spatial conditions for layout controllable generation. The disparity in conditioning inputs and supervision formats across datasets complicates their direct integration into a unified multi-task training framework. Third, existing unified models (e.g., Uni-RS) rely primarily on semantic guidance to steer generation. While such semantic representations effectively capture textual instructions, these representations ignore pixel-level spatial details, which limits the models’ ability to support generation tasks that require preserving the structure of the reference image, such as image inpainting and cross-modal transformation.



To address these challenges, we propose Omni-RS, a unified framework for multi-task remote sensing understanding and generation. As illustrated in Fig.~\ref{fig:intro}, Omni-RS harmonizes four understanding and five generation tasks within a single architecture. First, to mitigate cross-task interference caused by conflicting optimization objectives, we propose a ``Scene-Guided Two-Stage Training'' strategy. Motivated by the insight that both understanding and generation rely on a prior comprehension of geospatial scenes, this strategy decouples semantic understanding from visual synthesis into two stages. In the first stage, the MLLM is trained on remote sensing understanding tasks to construct a robust domain-specific scene representation. In the second stage, the learned hidden representations from the MLLM are projected into conditional embeddings for the diffusion generator, effectively leveraging the model's acquired scene awareness to guide image synthesis while mitigating cross-task interference. Second, to resolve dataset fragmentation, we construct UniGen, a unified remote sensing generation dataset that reorganizes heterogeneous data sources into a unified format. UniGen enables joint learning of multiple generation tasks across text-to-image synthesis, inpainting, outpainting, layout controllable generation, and cross-modal translation. Third, to overcome the loss of pixel-level spatial details in semantically guided generation, we introduce ``Detailed Visual Conditioning'', which injects VAE latent features from reference images into the diffusion generator. This mechanism supplies more visual cues, enabling the model to better preserve visible content.

In summary, our main contributions are as follows:

\begin{itemize}
    \item We propose \textbf{Omni-RS}, a unified remote sensing framework that harmonizes four understanding tasks and five generation tasks within a single architecture. Driven by a ``\textbf{Scene-Guided Two-Stage Training}'' strategy, Omni-RS leverages domain-specific scene understanding from an MLLM to effectively guide image generation.

    \item We construct \textbf{UniGen}, a unified remote sensing generation dataset that reorganizes different data sources into a unified format, enabling joint multi-task learning across text-to-image synthesis, inpainting, outpainting, layout control, and cross-modal translation.

    \item We introduce \textbf{Detailed Visual Conditioning}, a mechanism that injects VAE latents of reference images into the diffusion generator, effectively mitigating visual detail loss and ensuring high-fidelity structural preservation.
\end{itemize}





\section{Related Work}

\subsection{Remote Sensing Multimodal Understanding}

Early remote sensing understanding methods primarily relied on convolutional neural networks (CNNs) for task-specific visual interpretation, such as image classification and pixel-level segmentation~\citep{li2018deep,chauhan2018convolutional,wang2020comprehensive}. These approaches map remote sensing imagery to predefined categories or masks, providing effective solutions for individual perception tasks but offering limited support for flexible semantic interaction. With the development of large language models and vision-language learning, recent studies~\citep{zhang2023spatial,li2025segearth} have begun to integrate remote sensing imagery with language models, enabling users to interact with remote sensing scenes through natural-language instructions. For example, \citet{sun2025mask} enable models to describe changes in a scene using natural language, while \citet{zi2025rsvlm} require models to perform semantic, counting, and spatial reasoning in response to user queries.

\subsection{Remote Sensing Image Generation}

Early studies mainly employed generative adversarial networks (GANs) to synthesize remote sensing images~\citep{zhou2022towards,tao2023galip}. With the development of diffusion models, recent methods have substantially improved the fidelity and diversity of generated remote sensing images. For example, \citet{sebaq2024rsdiff} synthesize remote sensing scenes from natural-language descriptions, while \citet{tang2025aerogen} further incorporate layouts as spatial conditions to constrain scene structures. In addition, image-conditioned generation has also been explored for tasks. For example,~\citet{sun2023coarse} reconstruct missing regions conditioned on visible image content, while~\citet{li2022multiscale} transform imagery between different remote sensing modalities.

\subsection{Datasets for Remote Sensing Understanding And Generation}

The development of remote sensing multimodal understanding has benefited from increasingly mature data resources. Early datasets, such as RSICD, RSITMD, and RSVQA~\citep{lu2017exploring,yuan2022exploring,lobry2020rsvqa}, primarily support individual tasks such as image captioning and visual question answering. More recently, datasets such as VRSBench and GeoChat~\citep{li2024vrsbench,kuckreja2024geochat} have further integrated various understanding tasks into a unified format, laying the foundation for multimodal remote sensing understanding.

In contrast, existing remote sensing generation datasets are still primarily constructed for specific generation tasks. For example, Git-10M~\citep{liu2025text2earth} provides image and text pairs for text-to-image generation, while RSEdit~\citep{chen2026rsedit} pairs source images with editing instructions for image editing. TerraGen~\citep{tang2026terragen} further associates remote sensing images with layout-based spatial conditions to support controllable generation. However, these datasets differ substantially in their conditioning inputs and supervision formats, making them difficult to directly integrate for joint training across multiple generation tasks.

\subsection{Unified Understanding and Generation Models}

Recent studies have increasingly explored integrating multimodal understanding and generation within a single model. For example,~\citet{wang2024emu3} represent both images and text as discrete tokens and model them autoregressively within a shared sequence.~\citet{zhou2025transfusion} unify autoregressive language modeling and diffusion-based image generation within a single Transformer, jointly modeling discrete text tokens and continuous visual representations.~\citet{pan2025transfer} and~\citet{wu2026omnigen2} combine MLLMs with diffusion models, leveraging the semantic reasoning capability of the MLLM to provide conditions for high-fidelity image generation. More recently,~\citet{zhang2026uni} extend unified understanding and generation to remote sensing images. However, its generation capability remains limited to text-to-image generation and does not support more complex and practical image-conditioned tasks, such as image inpainting and modality translation. This limitation leaves the unified modeling of diverse remote sensing generation tasks largely unexplored.

\section{Dataset Pipeline}


\label{data}

\begin{figure*}[h]
  \centering
   \includegraphics[width=1\linewidth]{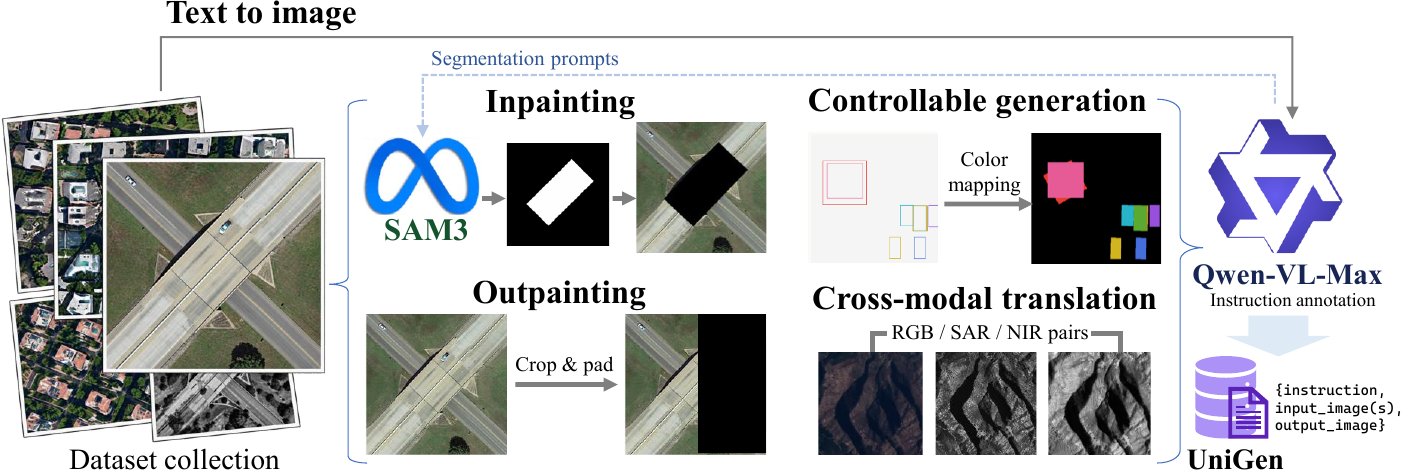}
   \caption{Overview of UniGen's construction pipeline. Heterogeneous remote sensing data are reorganized into five generation tasks, with Qwen-VL-Max providing captions and task instructions. The dashed arrow indicates the additional segmentation prompts supplied to SAM3 for inpainting.}
   \label{fig:pipline}
    \vspace{-1pt}
\end{figure*}

Existing remote sensing generation datasets are typically tailored for isolated tasks and seldom provide multiple generation tasks for the same geospatial scene. For instance, text-to-image datasets pair imagery with textual descriptions but they lack the reference images or task-specific instructions required for image inpainting. Likewise, datasets with spatial annotations or paired cross-modal observations provide disparate conditions that are not aligned across different tasks. To bridge this gap, we construct \textbf{UniGen}, a unified remote sensing dataset encompassing text-to-image, image inpainting, outpainting, controllable generation, and cross-modal translation. As illustrated in Figure~\ref{fig:pipline}, UniGen pairs task-specific reference images with tailored instruction annotations, standardizing heterogeneous data sources into a unified triplet format $\{\text{instruction}, \text{reference\_image}, \text{output\_image}\}$.

To build a comprehensive benchmark, we aggregate complementary data sources including Git-10M~\citep{liu2025text2earth}, DIOR-R~\citep{cheng2022anchor}, and WHU-OPT-SAR~\citep{li2022mcanet}. Specifically, Git-10M offers over 10 million globally distributed image–text pairs, providing diverse visual content and rich textual descriptions. DIOR-R contributes high-quality oriented bounding boxes and rotated object annotations, serving as a solid foundation for constructing precise spatial control conditions. Meanwhile, WHU-OPT-SAR supplies aligned RGB, NIR, and SAR observations at a 5~m spatial resolution across varied terrain, supporting cross-modal supervision for identical geographic regions. Leveraging these data sources, we derive task-specific instructions and reference image conditions around shared geospatial scenes to enable multi-task generation.

The construction pipeline of the UniGen dataset is illustrated in Figure~\ref{fig:pipline}. For text-to-image generation, we pair scene captions refined by Qwen-VL-Max~\citep{bai2023qwen} with their corresponding images. For inpainting, Qwen-VL-Max generates subject-specific restoration instructions and segmentation prompts, which steer SAM3~\citep{carion2026sam} to extract region masks; after filtering invalid masks, selected regions are removed to produce masked inputs while preserving original images as completion targets. For outpainting, we crop and pad images to construct partially observed canvases, pairing them with extension instructions and ground truth images. For controllable generation, to unify layout conditions with image conditions, we convert discrete annotations into rasterized color control maps. These control maps are generated by rendering DIOR-R's bounding boxes or by coloring category masks from the inpainting task. The task instructions are then generated by Qwen-VL-Max. For cross-modal translation, RGB inputs are paired with corresponding NIR or SAR targets under modality-specific instructions. For all tasks, Qwen-VL-Max generates multi-task instructions with a consistent style based on the target image, control conditions, and task category. Detailed source statistics are provided in
Appendix~\ref{app:unigen}.

\section{Method}


\begin{figure*}[h]
  \centering
  \includegraphics[width=1\linewidth]{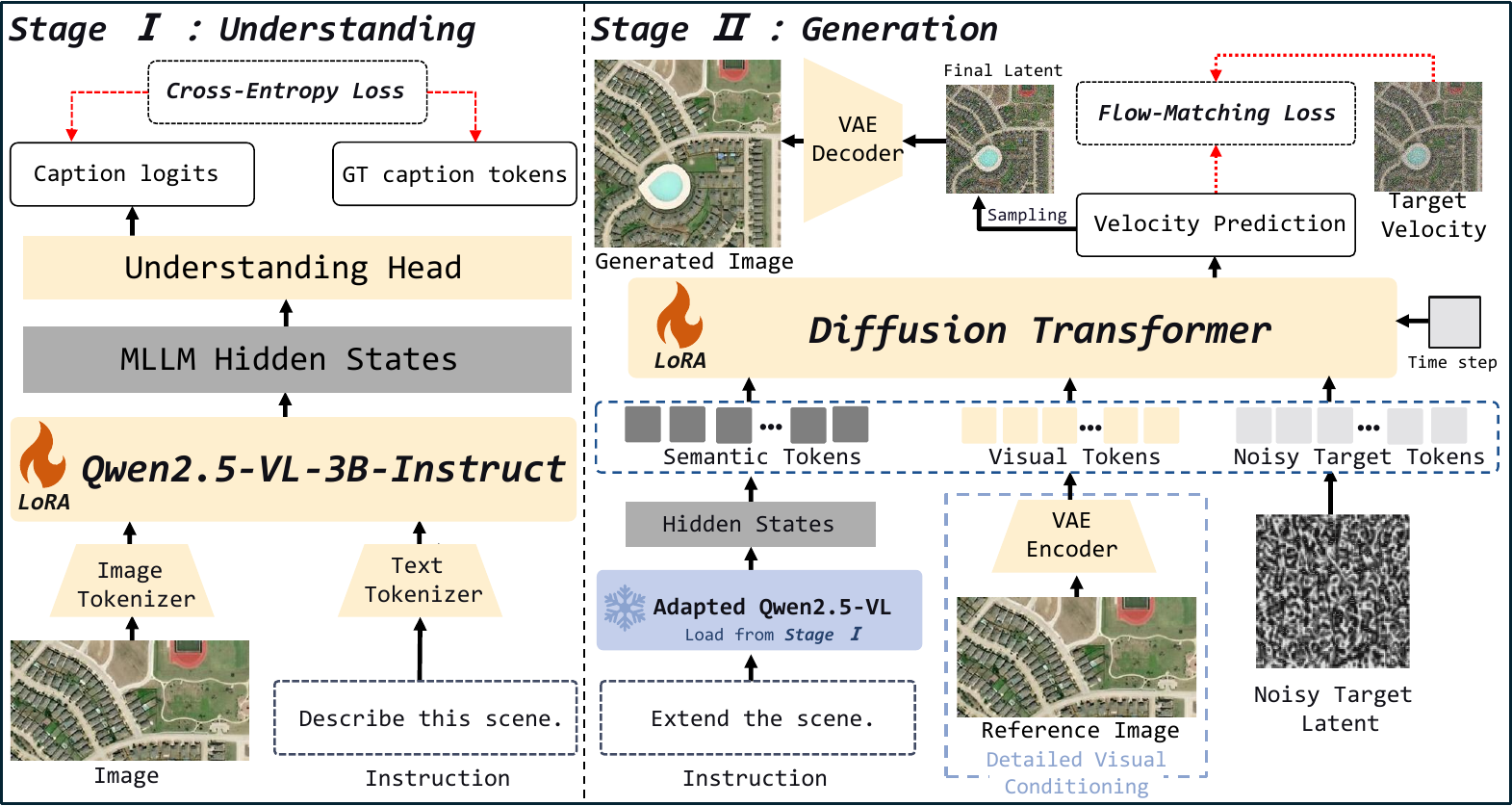}
  \caption{\textbf{Overview of Omni-RS.} Stage I adapts the MLLM to remote sensing understanding through LoRA tuning with a cross-entropy loss. Stage II freezes the adapted MLLM and trains the diffusion transformer through LoRA tuning with a flow-matching loss. Semantic tokens from the MLLM and visual tokens from ``Detailed Visual Conditioning'' jointly guide image generation.
}
  \label{fig:method}
\end{figure*}

\subsection{Overview}
\label{sec:overview}

As illustrated in Figure~\ref{fig:method}, Omni-RS bridges a multimodal large language model (MLLM) with a diffusion transformer (DiT) via hidden states, harmonizing remote sensing visual understanding and multi-task image generation within a unified architecture. Built upon Qwen2.5-VL-3B-Instruct~\citep{bai2025qwen25vltechnicalreport}, the MLLM interprets visual-textual inputs for understanding tasks while simultaneously serving as a semantic encoder for downstream generation. Formally, we represent a unified data sample as a tuple $(u, r, y)$, where $u$ denotes the text instruction, $r$ represents an optional reference image, and $y$ is the target output. For visual understanding, the MLLM processes the input image alongside instruction $u$ to autoregressively generate textual outputs. For generative tasks, the instruction $u$ and optional reference image $r$ are translated into unified semantic and visual tokens, respectively, which directly condition the DiT backbone to synthesize the target image $y$.


\subsection{Scene-Guided Two-Stage Training}
\label{sec:two_stage_training}

We observe that both remote sensing visual understanding and image generation rely on prior geographic scene modeling to capture high-level semantics and spatial topology before performing their respective tasks. Based on this key insight, we propose a novel ``\textbf{Scene-Guided Two-Stage Training}'' strategy. Specifically, this approach first adapts the MLLM to acquire domain-specific remote sensing knowledge, and subsequently leverages its frozen hidden states as conditional guidance to steer the DiT for multi-task image generation.

\paragraph{Stage I: Understanding Adaptation.}
We initialize the MLLM backbone with Qwen2.5-VL-3B-Instruct and fine-tune it on remote sensing understanding benchmarks. Given an input image $I$, an instruction prompt $q$, and a target textual sequence $a = (a_1, \dots, a_L)$, the model is optimized via an autoregressive cross-entropy loss:
\begin{equation}
\mathcal{L}_{\mathrm{under}} = -\mathbb{E}_{(I, q, a)} \left[ \sum_{j=1}^{L} \log p_{\phi}(a_j \mid I, q, a_{<j}) \right],
\end{equation}
where $\phi$ denotes the trainable Low-Rank Adaptation (LoRA) parameters. During this stage, the base MLLM backbone and the Diffusion Transformer remain frozen. Through supervision on dense scene descriptions, object categorizations, and spatial reasoning, the MLLM effectively aligns its semantic embedding space with remote sensing domain characteristics.

\paragraph{Stage II: Generation Training.}
In the second stage, we freeze the MLLM in Stage I and optimize the DiT, which is initialized from the pretrained OmniGen2~\citep{wu2026omnigen2}, using parameter-efficient LoRA adapters on the proposed UniGen dataset. Given a text instruction $u$, the adapted MLLM extracts multi-layer hidden states, which are mapped into semantic condition tokens $S$. Crucially, these tokens not only accurately convey the intent of the instruction $u$, but also inject the geospatial prior knowledge acquired in Stage I into the DiT. For tasks requiring reference images, $S$ is augmented with visual tokens $V$ derived via the Detailed Visual Conditioning module.


We train the generative backbone using a flow matching objective. Let $z = E(y)$ denote the target latent code extracted by a pretrained VAE encoder $E$, and let $\epsilon \sim \mathcal{N}(\mathbf{0}, \mathbf{I})$ represent standard Gaussian noise. At timestep $t \in [0, 1]$, the interpolated noisy latent is defined as $z_t = (1-t)\epsilon + tz$, with the target velocity given by $v^{*} = z - \epsilon$. Conditioned on semantic tokens $S$ and optional visual tokens $V$, the parameter-updated DiT $v_{\theta}$ is trained by minimizing the velocity regression loss:
\begin{equation}
\mathcal{L}_{\mathrm{gen}} = \mathbb{E}_{(u, r, y), t, \epsilon} \left[ \left\| v_{\theta}(z_t, t; S, V) - v^{*} \right\|_2^2 \right],
\end{equation}
where $\theta$ represents the trainable parameters of the DiT. This unified objective is applied across all five generative tasks in UniGen, with visual tokens $V$ set to empty ($\varnothing$) for pure text-to-image synthesis. During inference, the predicted velocity guides the sampling trajectory from pure Gaussian noise, and the final latent is projected back to image space via the VAE decoder. By decoupling semantic knowledge acquisition from visual generation, this two-stage paradigm enables the DiT to focus on high-fidelity visual synthesis conditioned on a robust, domain-adapted semantic foundation.

\subsection{Detailed Visual Conditioning}
\label{sec:detailed_visual_conditioning}

While semantic tokens capture high-level task intent, they inherently struggle to preserve the fine-grained visual details of reference images. Generative tasks in remote sensing, such as image inpainting and modality translation, require strict adherence to the visual context and structural constraints provided by a source image. To address this limitation, we propose Detailed Visual Conditioning to explicitly inject reference visual priors into the generative backbone.

Given a reference image $r$, a pretrained VAE encoder extracts its dense latent representation, which is subsequently tokenized and projected into visual tokens:
\begin{equation}
V = P_v(VAE(r)),
\end{equation}
where $P_v$ denotes the visual tokenization and linear projection layer. This pathway bypasses lossy textual compression, directly supplying spatially organized visual tokens alongside instruction-derived semantic tokens $S$. 

To integrate these multimodal signals, we construct a unified 1D token sequence by concatenating semantic tokens $S$, visual tokens $V$, and noisy target tokens $Z_t$ along the sequence dimension:
\begin{equation}
X = [S \,;\, V \,;\, Z_t] \in \mathbb{R}^{(L_s + L_v + L_z) \times D},
\end{equation}
where $[\cdot \,;\, \cdot]$ denotes sequence concatenation, $L_{(\cdot)}$ represents the corresponding sequence length, and $D$ is the embedding dimension. Fed with $X$ and timestep $t$, the DiT blocks model all cross-token interactions via full self-attention to predict the target velocity $v_\theta(X, t)$.

This conditioning pathway accommodates diverse visual inputs across tasks in a unified interface. Specifically, the reference image $r$ defined in Section~\ref{sec:overview} adapts to specific task requirements. For image inpainting and outpainting, $r$ provides the unmasked context; for controllable generation, $r$ introduces color maps for spatial layout positions; for cross-modal translation, $r$ supplies structural configurations from the source modality to ensure geometric consistency in the target output.

\section{Experiments}

\subsection{Datasets}

For training, we use GeoChat~\citep{kuckreja2024geochat} to adapt the MLLM for remote sensing understanding in Stage I and UniGen to train the generation branch in Stage II. Furthermore, we evaluate visual understanding on RSVQA-LR~\citep{lobry2020rsvqa}, VRSBench-VQA~\citep{li2024vrsbench}, and the GeoChat region understanding benchmark~\citep{kuckreja2024geochat}. For image generation, text-to-image generation is evaluated on RSICD~\citep{lu2017exploring}, layout-controllable generation on DIOR-R~\citep{cheng2022anchor}, and image completion (inpainting, outpainting) along with cross-modal translation (RGB$\rightarrow$NIR, RGB$\rightarrow$SAR) on the test split of our proposed \textbf{UniGen} dataset. Detailed descriptions and formulations of all evaluation metrics are provided in Appendix~\ref{sec:appendix_metrics}.



\subsection{Implementation Details}
Following Section~\ref{sec:two_stage_training}, we fine-tune the MLLM with rank-16 LoRA for 152K steps at a learning rate of $1\times10^{-4}$ in Stage I. In Stage II, we freeze the MLLM adapted in Stage I and train the diffusion transformer with rank-128 LoRA for 100K steps, using learning rates of $1\times10^{-4}$ for the first 55K steps and $5\times10^{-5}$ thereafter. Both stages are trained on two NVIDIA A100 GPUs.




\subsection{Comparison with Existing Methods}

\subsubsection{Remote Sensing Multimodal Understanding}

\begin{figure*}[h]
  \centering
   \includegraphics[width=0.85\linewidth]{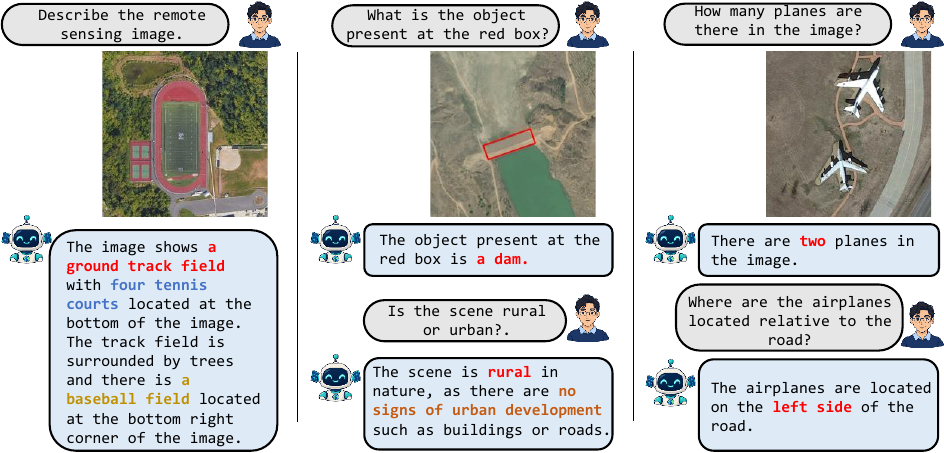}
   \caption{Qualitative results of Omni-RS on remote sensing understanding tasks.}
   \label{fig:vis_understanding}
\end{figure*}

\vspace{-10pt}

\begin{table}[h]
    \centering
    \caption{Comparison of Omni-RS with other methods on the RSVQA-LR dataset. The best results are shown in bold, and the second-best result is underlined. $\dagger$ denotes zero-shot evaluation.}
    \label{tab:rsvqa_lr}
    \setlength{\tabcolsep}{3.5pt}
    \begin{tabular}{lcccc}
        \toprule
        Method & Presence $\uparrow$ & Comparison $\uparrow$
        & Rural/Urban $\uparrow$ & Ave. $\uparrow$ \\
        \midrule
        RSVQA~\citep{lobry2020rsvqa}      & 87.46 & 81.50 & 90.00 & 84.64 \\
        EasyToHard~\citep{yuan2022easy} & 90.66 & 87.49 & 91.67 & 88.89 \\
        VMGN~\citep{liu2026visual}       & 91.30 & 92.52 & \textbf{94.00} & 92.03 \\
        RSUniVLM~\citep{liu2026rsunivlm}   & 91.51 & 92.65 & \underline{92.00} & 92.16 \\
        ScoreRS$^\dagger$~\citep{muhtar2026quality}    & 89.79 & 89.68 & 91.36 & 89.75 \\
        UniRS$^\dagger$~\citep{li2024unirs}      & \textbf{91.64} & \underline{92.68} & 90.00 & \underline{92.21} \\
        \midrule
        \textbf{Ours}$^\dagger$ & \underline{91.54} & \textbf{93.25} & \underline{92.00} & \textbf{92.52} \\
        \bottomrule
    \end{tabular}
\end{table}

\noindent\textbf{Qualitative Visual Comprehension.} 
Figure~\ref{fig:vis_understanding} showcases the visual comprehension of \mbox{Omni-RS} across varying ground scales and semantic granularities. In image captioning (left column), our model not only generates holistic scene narratives but also captures the spatial topology, such as the central runway surrounded by trees. In region-level perception (middle column), \mbox{Omni-RS} focuses strictly on the prompt-specified bounding box, accurately identifying the narrow infrastructure as a \textit{dam} and classifying the overarching scene as rural. Moreover, the right column highlights its capacity for fine-grained object enumeration (two aircraft) and relative spatial reasoning (localizing the planes to the \textit{left side} of the road). These results verify that our Stage~I adaptation successfully injects robust spatial awareness and geospatial priors into the MLLM backbone.

\vspace{-10pt}

\begin{table}[h]
    \centering
    \begin{minipage}[t]{0.49\textwidth}
        \vspace{0pt}
        \centering
        \caption{VQA results on VRSBench.}
        \label{tab:vrsbench_vqa}
        \small
        \renewcommand{\arraystretch}{1.0}
        \setlength{\tabcolsep}{2pt}
        \begin{tabular}{@{}lc@{}}
            \toprule
            Method & Accuracy $\uparrow$ \\
            \midrule
            ZoomEarth~\citep{liu2026zoomearth}
            & 37.50 \\
            SkySenseGPT~\citep{luo2024skysensegpt}
            & 39.80 \\
            GeoChat-7B~\citep{kuckreja2024geochat}
            & 46.54 \\
            InternVL-2.5~\citep{chen2024expanding}
            & 47.20 \\
            GeoLLaVA-8K~\citep{wang2026geollava}
            & 48.90 \\
            \midrule
            \textbf{Ours} & \textbf{49.89} \\
            \bottomrule
        \end{tabular}
    \end{minipage}
    \hfill
    \begin{minipage}[t]{0.49\textwidth}
        \vspace{0pt}
        \centering
        \caption{Text-to-image results on RSICD.}
        \label{tab:rsicd_t2i}
        \small
        \renewcommand{\arraystretch}{1.0}
        \setlength{\tabcolsep}{2pt}
        \begin{tabular}{@{}lcc@{}}
            \toprule
            Method & FID $\downarrow$ & CLIP $\uparrow$ \\
            \midrule
            DF-GAN~\citep{tao2022df}
            & 109.41 & 19.76 \\
            Txt2Img-MHN~\citep{xu2023txt2img}
            & 102.44 & 20.27 \\
            CRS-Diff~\citep{tang2024crs}
            & 50.72 & 20.33 \\
            LAFITE~\citep{zhou2022towards}
            & 74.11 & 22.52 \\
            Uni-RS~\citep{zhang2026uni}
            & \underline{22.11} & \textbf{25.68} \\
            \midrule
            \textbf{Ours}
            & \textbf{21.90} & \underline{24.85} \\
            \bottomrule
        \end{tabular}
    \end{minipage}
\end{table}

\noindent\textbf{Quantitative Evaluation.} 
Tables~\ref{tab:rsvqa_lr}, \ref{tab:vrsbench_vqa}, and \ref{tab:region_understanding} quantitatively evaluate the visual understanding capabilities of \mbox{Omni-RS} across diverse benchmarks. Under zero-shot evaluation on RSVQA-LR (Table~\ref{tab:rsvqa_lr}), \mbox{Omni-RS} achieves a state-of-the-art average accuracy of $92.52\%$, consistently outperforming existing specialized remote sensing methods and multimodal baselines. On VRSBench (Table~\ref{tab:vrsbench_vqa}), our 3B-parameter model obtains $49.89\%$ accuracy, noticeably surpassing substantially larger models such as GeoChat-7B ($+3.35\%$) and GeoLLaVA-8K ($+0.99\%$), demonstrating competitive parameter efficiency and effective domain adaptation. In the fine-grained region understanding benchmark (Table~\ref{tab:region_understanding}), although \mbox{Omni-RS} slightly trails the in-domain supervised GeoChat-7B, it substantially outperforms general-domain baselines like MiniGPT-v2 by a wide margin (e.g., $85.3$ vs. $32.1$ in ROUGE-1), confirming its solid region-level localization and comprehension capabilities.


\vspace{-10pt}

\begin{table}[h]
    \centering
    \begin{minipage}[t]{0.49\textwidth}
        \vspace{0pt}
        \centering
        \caption{Region understanding on GeoChat.}
        \label{tab:region_understanding}
        \small
        \renewcommand{\arraystretch}{1.0}
        \setlength{\tabcolsep}{2pt}
        \begin{tabular}{@{}lccc@{}}
            \toprule
            Method & R-1 $\uparrow$ & R-L $\uparrow$
            & MR $\uparrow$ \\
            \midrule
            MiniGPT-v2~\citep{chen2023minigpt}
            & 32.1 & 31.2 & 10.0 \\
            GeoChat~\citep{kuckreja2024geochat}
            & \textbf{87.3} & \textbf{87.2}
            & \textbf{83.9} \\
            \midrule
            \textbf{Ours}
            & \underline{85.3} & \underline{85.3}
            & \underline{80.6} \\
            \bottomrule
        \end{tabular}
    \end{minipage}
    \hfill
    \begin{minipage}[t]{0.49\textwidth}
        \vspace{0pt}
        \centering
        \caption{Controllable generation on DIOR-R.}
        \label{tab:control}
        \small
        \renewcommand{\arraystretch}{1.0}
        \setlength{\tabcolsep}{2pt}
        \begin{tabular}{@{}lcc@{}}
            \toprule
            Method & YOLO Score $\uparrow$ & CAS $\uparrow$ \\
            \midrule
            GLIGEN~\citep{li2023gligen}
            & 24.6/41.6/25.1 & 58.89 \\
            AeroGen~\citep{tang2025aerogen}
            & \underline{29.6}/\textbf{57.6}/\underline{32.0}
            & \underline{74.13} \\
            \midrule
            \textbf{Ours}
            & \textbf{31.4}/\underline{47.8}/\textbf{32.1}
            & \textbf{78.77} \\
            \bottomrule
        \end{tabular}
    \end{minipage}
\end{table}

\vspace{-15pt}

\begin{table}[h]
    \centering
    \caption{Comparison of cross-modal remote sensing image translation methods. 
    }
    \label{tab:multimodal}
    \setlength{\tabcolsep}{3.5pt}
    \begin{tabular}{lccc|ccc}
        \toprule
        & \multicolumn{3}{c|}{RGB $\rightarrow$ NIR}
        & \multicolumn{3}{c}{RGB $\rightarrow$ SAR} \\
        Method
        & MAE $\downarrow$ & SSIM $\uparrow$ & PSNR $\uparrow$
        & MAE $\downarrow$ & RMSE $\downarrow$ & PSNR $\uparrow$ \\
        \midrule
        pix2pix~\citep{isola2017image}
        & 0.64 & 0.126 & 5.27
        & 0.52 & 0.67 & 3.47 \\
        Text2Earth~\citep{liu2025text2earth}
        & 0.19 & 0.166 & 12.96
        & 0.28 & 0.31 & 10.23 \\
        RSEdit~\citep{chen2026rsedit}
        & 0.30 & 0.097 & 9.24
        & 0.29 & 0.36 & 8.99 \\
        \midrule
        \textbf{Ours}
        & \textbf{0.09} & \textbf{0.655} & \textbf{18.68}
        & \textbf{0.17} & \textbf{0.23} & \textbf{12.89} \\
        \bottomrule
    \end{tabular}
\end{table}
\subsubsection{Multi-Task Remote Sensing Generation}

\begin{figure*}[t]
  \centering
   \includegraphics[width=0.85\linewidth]{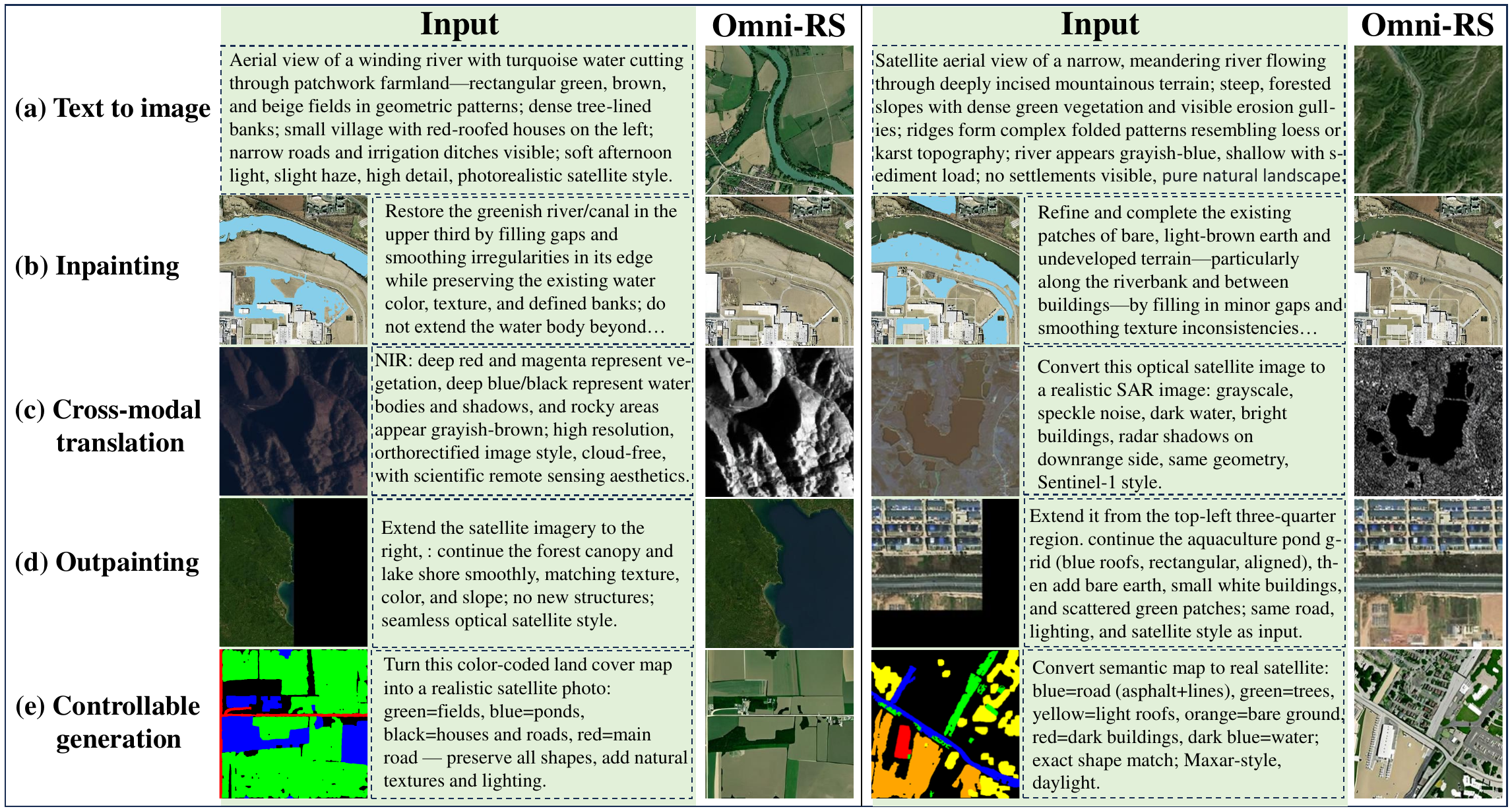}
   \caption{Qualitative results of Omni-RS on remote sensing generation tasks.}
   \label{fig:vis_generation}
   \vspace{-10pt}
\end{figure*}

\noindent\textbf{Qualitative Generation Fidelity.} 
Figure~\ref{fig:vis_generation} visually illustrates the multi-task generative capabilities of \mbox{Omni-RS} across diverse scenarios within a single unified architecture. Beyond text-to-image synthesis, for inpainting and outpainting, \mbox{Omni-RS} enables continuous landscape features, such as waterways and road grids, to extend naturally across mask boundaries. Furthermore, in cross-modal translation and controllable generation, the model reliably preserves structural topologies while simultaneously accommodating sensor-specific physical properties and categorical spatial constraints. Additional comparisons and large-scale outpainting results are detailed in Appendix~\ref{app:qualitative} and~\ref{app:outpainting}.


\noindent\textbf{Quantitative Evaluation across Generative Tasks.} 
The quantitative results in Tables~\ref{tab:rsicd_t2i}, \ref{tab:control}, \ref{tab:multimodal}, and \ref{tab:inpaint_outpaint} validate the effectiveness and structural fidelity of \mbox{Omni-RS} across varied synthesis tasks. 
In text-to-image generation (Table~\ref{tab:rsicd_t2i}), \mbox{Omni-RS} attains the lowest FID of $21.90$ alongside a competitive CLIP score ($24.85$) on RSICD. These results demonstrate that incorporating reference-conditioned generation tasks via Detailed Visual Conditioning does not compromise the text-to-image synthesis capability. 
In layout-controllable generation (Table~\ref{tab:control}) and cross-modal translation (Table~\ref{tab:multimodal}), \mbox{Omni-RS} excels at maintaining spatial layouts and geometric structures. Specifically, the leading Classification Accuracy Score ($78.77$ CAS) on DIOR-R verifies strict adherence to spatial bounding constraints, while the superior PSNR in RGB to NIR/SAR translation confirms that our visual conditioning pathway robustly preserves macroscopic ground geometry across distinct sensor domains. 
Furthermore, for inpainting and outpainting (Table~\ref{tab:inpaint_outpaint}), our framework achieves significant performance gains in both inpainting (reducing LPIPS by $0.12$ compared to the second-best baseline) and outpainting ($35.33$ FID), confirming that our injection of VAE latent tokens allows for the preservation of unmasked regions and ensures seamless continuity at the boundaries.


\vspace{-10pt}
\begin{table}[h]
    \centering
    \caption{Comparison of image inpainting and outpainting methods.}
    \label{tab:inpaint_outpaint}
    \setlength{\tabcolsep}{3pt}
    \begin{tabular}{lccc|ccc}
        \toprule
        & \multicolumn{3}{c|}{Inpainting} 
        & \multicolumn{3}{c}{Outpainting} \\
        Method & FID$\downarrow$ & KID$\downarrow$ & LPIPS$\downarrow$
        & FID$\downarrow$ & KID$\downarrow$ & LPIPS$\downarrow$ \\
        \midrule

        PowerPaint~\citep{zhuang2024task}  & 129.98 & 0.0614 & 0.46
                    & 67.84 & 0.0236 & \underline{0.24} \\
        SD2~\citep{rombach2022high}         & \underline{47.98} & \underline{0.0092} & \underline{0.24}
                    & 70.33 & 0.0224 & 0.32 \\
        RSEdit~\citep{chen2026rsedit} & 87.75 & 0.0304 & 0.53
                    & 81.22 & 0.0310 & 0.60 \\
        Text2Earth~\citep{liu2025text2earth}  & 63.67 & 0.0192 & 0.25
                    & \underline{66.65} & \underline{0.0180} & 0.27 \\

        \midrule
        \textbf{Ours} & \textbf{31.23} & \textbf{0.0025} & \textbf{0.12}
        & \textbf{35.33} & \textbf{0.0018} & \textbf{0.22} \\
        \bottomrule
    \end{tabular}
\end{table}

\section{Conclusion and Future Work}

\label{sec:conclusion}

In this paper, we presented \textbf{Omni-RS}, a unified multimodal framework that seamlessly bridges remote sensing visual understanding and multi-task image generation within a single architecture. By harnessing an MLLM as the universal task planner and conditioning engine alongside a DiT backbone, Omni-RS effectively eliminates the barrier between perception and synthesis in Earth observation. Furthermore, our proposed two-stage training paradigm, complemented by the newly curated \textbf{UniGen} benchmark, establishes a strong baseline for complex tasks such as layout-controllable generation, high-fidelity inpainting/outpainting, and cross-modal RGB-to-SAR/NIR translation. Extensive quantitative and qualitative evaluations demonstrate that Omni-RS achieves state-of-the-art or competitive performance across different benchmark tasks.

\paragraph{Limitations and Future Work.}

Although Omni-RS unifies multiple tasks in remote sensing understanding and generation, the model’s understanding task output is currently limited to discrete text sequences and lacks pixel-level visual understanding capabilities, such as image segmentation. In future research, we plan to leverage the powerful open-domain reasoning capabilities of MLLMs to extend the understanding task to open-vocabulary segmentation and reference segmentation. This will further advance the development of a comprehensive remote sensing model that unifies text, masks, and pixels within a single framework.

\bibliography{iclr2027_conference}

@misc{bai2025qwen25vltechnicalreport,
      title={Qwen2.5-VL Technical Report}, 
      author={Shuai Bai and Keqin Chen and Xuejing Liu and Jialin Wang and Wenbin Ge and Sibo Song and Kai Dang and Peng Wang and Shijie Wang and Jun Tang and Humen Zhong and Yuanzhi Zhu and Mingkun Yang and Zhaohai Li and Jianqiang Wan and Pengfei Wang and Wei Ding and Zheren Fu and Yiheng Xu and Jiabo Ye and Xi Zhang and Tianbao Xie and Zesen Cheng and Hang Zhang and Zhibo Yang and Haiyang Xu and Junyang Lin},
      year={2025},
      eprint={2502.13923},
      archivePrefix={arXiv},
      primaryClass={cs.CV},
      url={https://arxiv.org/abs/2502.13923}, 
}

@inproceedings{chauhan2018convolutional,
  title={Convolutional neural network (CNN) for image detection and recognition},
  author={Chauhan, Rahul and Ghanshala, Kamal Kumar and Joshi, Ramesh Chandra},
  booktitle={2018 first international conference on secure cyber computing and communication (ICSCCC)},
  pages={278--282},
  year={2018},
  organization={IEEE}
}

@article{maxwell2018implementation,
  title={Implementation of machine-learning classification in remote sensing: An applied review},
  author={Maxwell, Aaron E and Warner, Timothy A and Fang, Fang},
  journal={International journal of remote sensing},
  volume={39},
  number={9},
  pages={2784--2817},
  year={2018},
  publisher={Taylor \& Francis}
}

@article{van2013remote,
  title={Remote sensing and GIS for natural hazards assessment and disaster risk management},
  author={Van Westen, Cees J},
  journal={Treatise on geomorphology},
  volume={3},
  number={15},
  pages={259--298},
  year={2013},
  publisher={Academic Press San Diego, CA}
}

@article{li2020review,
  title={A review of remote sensing for environmental monitoring in China},
  author={Li, Jun and Pei, Yanqiu and Zhao, Shaohua and Xiao, Rulin and Sang, Xiao and Zhang, Chengye},
  journal={Remote Sensing},
  volume={12},
  number={7},
  pages={1130},
  year={2020},
  publisher={MDPI}
}

@article{wellmann2020remote,
  title={Remote sensing in urban planning: Contributions towards ecologically sound policies?},
  author={Wellmann, Thilo and Lausch, Angela and Andersson, Erik and Knapp, Sonja and Cortinovis, Chiara and Jache, Jessica and Scheuer, Sebastian and Kremer, Peleg and Mascarenhas, Andr{\'e} and Kraemer, Roland and others},
  journal={Landscape and urban planning},
  volume={204},
  pages={103921},
  year={2020},
  publisher={Elsevier}
}

@article{li2024vrsbench,
  title={Vrsbench: A versatile vision-language benchmark dataset for remote sensing image understanding},
  author={Li, Xiang and Ding, Jian and Elhoseiny, Mohamed},
  journal={Advances in Neural Information Processing Systems},
  volume={37},
  pages={3229--3242},
  year={2024}
}

@article{lobry2020rsvqa,
  title={RSVQA: Visual question answering for remote sensing data},
  author={Lobry, Sylvain and Marcos, Diego and Murray, Jesse and Tuia, Devis},
  journal={IEEE Transactions on Geoscience and Remote Sensing},
  volume={58},
  number={12},
  pages={8555--8566},
  year={2020},
  publisher={IEEE}
}

@article{tang2024crs,
  title={Crs-diff: Controllable remote sensing image generation with diffusion model},
  author={Tang, Datao and Cao, Xiangyong and Hou, Xingsong and Jiang, Zhongyuan and Liu, Junmin and Meng, Deyu},
  journal={IEEE Transactions on Geoscience and Remote Sensing},
  volume={62},
  pages={1--14},
  year={2024},
  publisher={IEEE}
}

@article{chen2026rsedit,
  title={RSEdit: Text-Guided Image Editing for Remote Sensing},
  author={Chen, Zhenyuan and Zhang, Zechuan and Zhang, Feng},
  journal={IEEE Geoscience and Remote Sensing Letters},
  year={2026},
  publisher={IEEE}
}

@article{yuan2022exploring,
  title={Exploring a fine-grained multiscale method for cross-modal remote sensing image retrieval},
  author={Yuan, Zhiqiang and Zhang, Wenkai and Fu, Kun and Li, Xuan and Deng, Chubo and Wang, Hongqi and Sun, Xian},
  journal={arXiv preprint arXiv:2204.09868},
  year={2022}
}

@article{li2022multiscale,
  title={Multiscale generative adversarial network based on wavelet feature learning for SAR-to-optical image translation},
  author={Li, Huihui and Gu, Cang and Wu, Dongqing and Cheng, Gong and Guo, Lei and Liu, Hang},
  journal={IEEE Transactions on Geoscience and Remote Sensing},
  volume={60},
  pages={1--15},
  year={2022},
  publisher={IEEE}
}

@article{sun2023coarse,
  title={Coarse-to-fine task-driven inpainting for geoscience images},
  author={Sun, Huiming and Ma, Jin and Guo, Qing and Zou, Qin and Song, Shaoyue and Lin, Yuewei and Yu, Hongkai},
  journal={IEEE Transactions on Circuits and Systems for Video Technology},
  volume={33},
  number={12},
  pages={7170--7182},
  year={2023},
  publisher={IEEE}
}

@inproceedings{li2023gligen,
  title={Gligen: Open-set grounded text-to-image generation},
  author={Li, Yuheng and Liu, Haotian and Wu, Qingyang and Mu, Fangzhou and Yang, Jianwei and Gao, Jianfeng and Li, Chunyuan and Lee, Yong Jae},
  booktitle={2023 IEEE/CVF Conference on Computer Vision and Pattern Recognition (CVPR)},
  pages={22511--22521},
  year={2023},
  organization={IEEE}
}

@inproceedings{tao2022df,
  title={Df-gan: A simple and effective baseline for text-to-image synthesis},
  author={Tao, Ming and Tang, Hao and Wu, Fei and Jing, Xiao-Yuan and Bao, Bing-Kun and Xu, Changsheng},
  booktitle={Proceedings of the IEEE/CVF conference on computer vision and pattern recognition},
  pages={16515--16525},
  year={2022}
}

@inproceedings{zhou2022towards,
  title={Towards language-free training for text-to-image generation},
  author={Zhou, Yufan and Zhang, Ruiyi and Chen, Changyou and Li, Chunyuan and Tensmeyer, Chris and Yu, Tong and Gu, Jiuxiang and Xu, Jinhui and Sun, Tong},
  booktitle={2022 IEEE/CVF Conference on Computer Vision and Pattern Recognition (CVPR)},
  pages={17886--17896},
  year={2022},
  organization={IEEE}
}

@article{cheng2022anchor,
  title={Anchor-free oriented proposal generator for object detection},
  author={Cheng, Gong and Wang, Jiabao and Li, Ke and Xie, Xingxing and Lang, Chunbo and Yao, Yanqing and Han, Junwei},
  journal={IEEE Transactions on Geoscience and Remote Sensing},
  volume={60},
  pages={1--11},
  year={2022},
  publisher={IEEE}
}

@inproceedings{rombach2022high,
  title={High-resolution image synthesis with latent diffusion models},
  author={Rombach, Robin and Blattmann, Andreas and Lorenz, Dominik and Esser, Patrick and Ommer, Bj{\"o}rn},
  booktitle={2022 IEEE/CVF conference on computer vision and pattern recognition (CVPR)},
  pages={10674--10685},
  year={2022},
  organization={ieee}
}

@article{liu2025text2earth,
  title={Text2Earth: Unlocking text-driven remote sensing image generation with a global-scale dataset and a foundation model},
  author={Liu, Chenyang and Chen, Keyan and Zhao, Rui and Zou, Zhengxia and Shi, Zhenwei},
  journal={IEEE Geoscience and Remote Sensing Magazine},
  volume={13},
  number={3},
  pages={238--259},
  year={2025},
  publisher={IEEE}
}

@article{xu2023txt2img,
  title={Txt2Img-MHN: Remote sensing image generation from text using modern Hopfield networks},
  author={Xu, Yonghao and Yu, Weikang and Ghamisi, Pedram and Kopp, Michael and Hochreiter, Sepp},
  journal={IEEE Transactions on Image Processing},
  volume={32},
  pages={5737--5750},
  year={2023},
  publisher={IEEE}
}

@article{sebaq2024rsdiff,
  title={Rsdiff: Remote sensing image generation from text using diffusion model},
  author={Sebaq, Ahmad and ElHelw, Mohamed},
  journal={Neural Computing and Applications},
  volume={36},
  number={36},
  pages={23103--23111},
  year={2024},
  publisher={Springer}
}

@inproceedings{zhuang2024task,
  title={A task is worth one word: Learning with task prompts for high-quality versatile image inpainting},
  author={Zhuang, Junhao and Zeng, Yanhong and Liu, Wenran and Yuan, Chun and Chen, Kai},
  booktitle={European Conference on Computer Vision},
  pages={195--211},
  year={2024},
  organization={Springer}
}

@article{chen2023minigpt,
  title={Minigpt-v2: large language model as a unified interface for vision-language multi-task learning},
  author={Chen, Jun and Zhu, Deyao and Shen, Xiaoqian and Li, Xiang and Liu, Zechun and Zhang, Pengchuan and Krishnamoorthi, Raghuraman and Chandra, Vikas and Xiong, Yunyang and Elhoseiny, Mohamed},
  journal={arXiv preprint arXiv:2310.09478},
  year={2023}
}

@article{wang2026geollava,
  title={Geollava-8k: Scaling remote-sensing multimodal large language models to 8k resolution},
  author={Wang, Fengxiang and Chen, Mingshuo and Li, Yueying and Wang, Di and Wang, Haotian and Guo, Zonghao and Wang, Zefan and Boqi, Shan and Lan, Long and Wang, Yulin and others},
  journal={Advances in Neural Information Processing Systems},
  volume={38},
  pages={159185--159218},
  year={2026}
}

@inproceedings{liu2026zoomearth,
  title={Zoomearth: Active perception for ultra-high-resolution geospatial vision-language tasks},
  author={Liu, Ruixun and Fu, Bowen and Song, Jiayi and Li, Kaiyu and Li, Wanchen and Xue, Lanxuan and Qiao, Hui and Zhang, Weizhan and Meng, Deyu and Cao, Xiangyong},
  booktitle={Proceedings of the IEEE/CVF Conference on Computer Vision and Pattern Recognition},
  pages={34877--34888},
  year={2026}
}

@article{chen2024expanding,
  title={Expanding performance boundaries of open-source multimodal models with model, data, and test-time scaling},
  author={Chen, Zhe and Wang, Weiyun and Cao, Yue and Liu, Yangzhou and Gao, Zhangwei and Cui, Erfei and Zhu, Jinguo and Ye, Shenglong and Tian, Hao and Liu, Zhaoyang and others},
  journal={arXiv preprint arXiv:2412.05271},
  year={2024}
}

@article{li2024unirs,
  title={UniRS: Unifying multi-temporal remote sensing tasks through vision language models},
  author={Li, Yujie and Xu, Wenjia and Li, Guangzuo and Yu, Zijian and Wei, Zhiwei and Wang, Jiuniu and Peng, Mugen},
  journal={arXiv preprint arXiv:2412.20742},
  year={2024}
}

@article{liu2026rsunivlm,
  title={RSUniVLM: A unified vision-language model for remote sensing via Granularity-oriented MoE},
  author={Liu, Xu and Lian, Zhouhui},
  journal={Pattern Recognition},
  pages={113717},
  year={2026},
  publisher={Elsevier}
}

@article{liu2026visual,
  title={A visual-textual mutual guidance fusion network for remote sensing visual question answering},
  author={Liu, Haolin and Chen, Lei and Lu, Xinchao and Wang, Hao and Bai, Lu and Wang, Maoli and Ren, Peng},
  journal={Pattern Recognition},
  volume={176},
  pages={113258},
  year={2026},
  publisher={Elsevier}
}

@article{muhtar2026quality,
  title={Quality-driven curation of remote sensing vision-language data via learned scoring models},
  author={Muhtar, Dilxat and Zhang, Enzhuo and Li, Zhenshi and Gu, Feng and He, Yanglangxing and Xiao, Pengfeng and Zhang, Xueliang},
  journal={Advances in Neural Information Processing Systems},
  volume={38},
  pages={171088--171133},
  year={2026}
}

@article{yuan2022easy,
  title={From easy to hard: Learning language-guided curriculum for visual question answering on remote sensing data},
  author={Yuan, Zhenghang and Mou, Lichao and Wang, Qi and Zhu, Xiao Xiang},
  journal={IEEE transactions on geoscience and remote sensing},
  volume={60},
  pages={1--11},
  year={2022},
  publisher={IEEE}
}

@article{wang2004image,
  title={Image quality assessment: from error visibility to structural similarity},
  author={Wang, Zhou and Bovik, Alan C and Sheikh, Hamid R and Simoncelli, Eero P},
  journal={IEEE transactions on image processing},
  volume={13},
  number={4},
  pages={600--612},
  year={2004},
  publisher={IEEE}
}

@article{ravuri2019classification,
  title={Classification accuracy score for conditional generative models},
  author={Ravuri, Suman and Vinyals, Oriol},
  journal={Advances in neural information processing systems},
  volume={32},
  year={2019}
}

@inproceedings{li2021image,
  title={Image synthesis from layout with locality-aware mask adaption},
  author={Li, Zejian and Wu, Jingyu and Koh, Immanuel and Tang, Yongchuan and Sun, Lingyun},
  booktitle={2021 IEEE/CVF International Conference on Computer Vision (ICCV)},
  pages={13799--13808},
  year={2021},
  organization={IEEE}
}

@inproceedings{zhang2018unreasonable,
  title={The unreasonable effectiveness of deep features as a perceptual metric},
  author={Zhang, Richard and Isola, Phillip and Efros, Alexei A and Shechtman, Eli and Wang, Oliver},
  booktitle={2018 IEEE/CVF conference on computer vision and pattern recognition},
  pages={586--595},
  year={2018},
  organization={IEEE}
}

@article{binkowski2018demystifying,
  title={Demystifying mmd gans},
  author={Bi{\'n}kowski, Miko{\l}aj and Sutherland, Danica J and Arbel, Michael and Gretton, Arthur},
  journal={arXiv preprint arXiv:1801.01401},
  year={2018}
}

@inproceedings{radford2021learning,
  title={Learning transferable visual models from natural language supervision},
  author={Radford, Alec and Kim, Jong Wook and Hallacy, Chris and Ramesh, Aditya and Goh, Gabriel and Agarwal, Sandhini and Sastry, Girish and Askell, Amanda and Mishkin, Pamela and Clark, Jack and others},
  booktitle={International conference on machine learning},
  pages={8748--8763},
  year={2021},
  organization={PmLR}
}

@article{heusel2017gans,
  title={Gans trained by a two time-scale update rule converge to a local nash equilibrium},
  author={Heusel, Martin and Ramsauer, Hubert and Unterthiner, Thomas and Nessler, Bernhard and Hochreiter, Sepp},
  journal={Advances in neural information processing systems},
  volume={30},
  year={2017}
}

@inproceedings{banerjee2005meteor,
  title={METEOR: An automatic metric for MT evaluation with improved correlation with human judgments},
  author={Banerjee, Satanjeev and Lavie, Alon},
  booktitle={Proceedings of the acl workshop on intrinsic and extrinsic evaluation measures for machine translation and/or summarization},
  pages={65--72},
  year={2005}
}

@inproceedings{lin2004rouge,
  title={Rouge: A package for automatic evaluation of summaries},
  author={Lin, Chin-Yew},
  booktitle={Text summarization branches out},
  pages={74--81},
  year={2004}
}

@inproceedings{isola2017image,
  title={Image-to-image translation with conditional adversarial networks},
  author={Isola, Phillip and Zhu, Jun-Yan and Zhou, Tinghui and Efros, Alexei A},
  booktitle={2017 IEEE conference on computer vision and pattern recognition (CVPR)},
  pages={5967--5976},
  year={2017},
  organization={Ieee}
}

@article{lu2017exploring,
  title={Exploring models and data for remote sensing image caption generation},
  author={Lu, Xiaoqiang and Wang, Binqiang and Zheng, Xiangtao and Li, Xuelong},
  journal={IEEE Transactions on Geoscience and Remote Sensing},
  volume={56},
  number={4},
  pages={2183--2195},
  year={2017},
  publisher={IEEE}
}

@inproceedings{carion2026sam,
  title={Sam 3: Segment anything with concepts},
  author={Carion, Nicolas and Gustafson, Laura and Hu, Yuan-Ting and Debnath, Shoubhik and Hu, Ronghang and Suris Coll-Vinent, Didac and Ryali, Chaitanya and Alwala, Kalyan Vasudev and Khedr, Haitham and Huang, Andrew and others},
  booktitle={International conference on learning representations},
  volume={2026},
  pages={138846--138923},
  year={2026}
}

@article{bai2023qwen,
  title={Qwen technical report},
  author={Bai, Jinze and Bai, Shuai and Chu, Yunfei and Cui, Zeyu and Dang, Kai and Deng, Xiaodong and Fan, Yang and Ge, Wenbin and Han, Yu and Huang, Fei and others},
  journal={arXiv preprint arXiv:2309.16609},
  year={2023}
}

@article{li2022mcanet,
  title={MCANet: A joint semantic segmentation framework of optical and SAR images for land use classification},
  author={Li, Xue and Zhang, Guo and Cui, Hao and Hou, Shasha and Wang, Shunyao and Li, Xin and Chen, Yujia and Li, Zhijiang and Zhang, Li},
  journal={International Journal of Applied Earth Observation and Geoinformation},
  volume={106},
  pages={102638},
  year={2022},
  publisher={Elsevier}
}

@inproceedings{wu2026omnigen2,
  title={Omnigen2: Towards instruction-aligned multimodal generation},
  author={Wu, Chenyuan and Wang, Jiahao and Zheng, Pengfei and Yan, Ruiran and Xiao, Shitao and Luo, Xin and Wang, Yueze and Li, Wanli and Jiang, Xiyan and Liu, Yexin and others},
  booktitle={Proceedings of the IEEE/CVF Conference on Computer Vision and Pattern Recognition},
  pages={21964--21975},
  year={2026}
}

@article{pan2025transfer,
  title={Transfer between modalities with metaqueries},
  author={Pan, Xichen and Shukla, Satya Narayan and Singh, Aashu and Zhao, Zhuokai and Mishra, Shlok Kumar and Wang, Jialiang and Xu, Zhiyang and Chen, Jiuhai and Li, Kunpeng and Juefei-Xu, Felix and others},
  journal={arXiv preprint arXiv:2504.06256},
  year={2025}
}

@inproceedings{zhou2025transfusion,
  title={Transfusion: Predict the next token and diffuse images with one multi-modal model},
  author={Zhou, Chunting and Yu, Lili and Babu, Arun and Tirumala, Kushal and Yasunaga, Michihiro and Shamis, Leonid and Kahn, Jacob and Ma, Xuezhe and Zettlemoyer, Luke and Levy, Omer},
  booktitle={International Conference on Learning Representations},
  volume={2025},
  pages={6446--6469},
  year={2025}
}

@article{wang2024emu3,
  title={Emu3: Next-token prediction is all you need},
  author={Wang, Xinlong and Zhang, Xiaosong and Luo, Zhengxiong and Sun, Quan and Cui, Yufeng and Wang, Jinsheng and Zhang, Fan and Wang, Yueze and Li, Zhen and Yu, Qiying and others},
  journal={arXiv preprint arXiv:2409.18869},
  year={2024}
}

@inproceedings{kuckreja2024geochat,
  title={Geochat: Grounded large vision-language model for remote sensing},
  author={Kuckreja, Kartik and Danish, Muhammad Sohail and Naseer, Muzammal and Das, Abhijit and Khan, Salman and Khan, Fahad Shahbaz},
  booktitle={2024 IEEE/CVF Conference on Computer Vision and Pattern Recognition (CVPR)},
  pages={27831--27840},
  year={2024},
  organization={IEEE}
}

@article{luo2024skysensegpt,
  title={Skysensegpt: A fine-grained instruction tuning dataset and model for remote sensing vision-language understanding},
  author={Luo, Junwei and Pang, Zhen and Zhang, Yongjun and Wang, Tingzhu and Wang, Linlin and Dang, Bo and Lao, Jiangwei and Wang, Jian and Chen, Jingdong and Tan, Yihua and others},
  journal={arXiv preprint arXiv:2406.10100},
  year={2024}
}

@inproceedings{tao2023galip,
  title={Galip: Generative adversarial clips for text-to-image synthesis},
  author={Tao, Ming and Bao, Bing-Kun and Tang, Hao and Xu, Changsheng},
  booktitle={2023 IEEE/CVF Conference on Computer Vision and Pattern Recognition (CVPR)},
  pages={14214--14223},
  year={2023},
  organization={IEEE}
}

@inproceedings{zi2025rsvlm,
  title={Rsvlm-qa: A benchmark dataset for remote sensing vision language model-based question answering},
  author={Zi, Xing and Xiao, Jinghao and Shi, Yunxiao and Tao, Xian and Li, Jun and Braytee, Ali and Prasad, Mukesh},
  booktitle={Proceedings of the 33rd ACM International Conference on Multimedia},
  pages={12905--12911},
  year={2025}
}

@article{sun2025mask,
  title={Mask approximation net: A novel diffusion model approach for remote sensing change captioning},
  author={Sun, Dongwei and Yao, Jing and Xue, Wu and Zhou, Changsheng and Ghamisi, Pedram and Cao, Xiangyong},
  journal={IEEE transactions on geoscience and remote sensing},
  volume={63},
  pages={1--11},
  year={2025},
  publisher={IEEE}
}

@inproceedings{li2025segearth,
  title={Segearth-ov: Towards training-free open-vocabulary segmentation for remote sensing images},
  author={Li, Kaiyu and Liu, Ruixun and Cao, Xiangyong and Bai, Xueru and Zhou, Feng and Meng, Deyu and Wang, Zhi},
  booktitle={2025 IEEE/CVF Conference on Computer Vision and Pattern Recognition (CVPR)},
  pages={10545--10556},
  year={2025},
  organization={IEEE}
}

@article{zhang2023spatial,
  title={A spatial hierarchical reasoning network for remote sensing visual question answering},
  author={Zhang, Zixiao and Jiao, Licheng and Li, Lingling and Liu, Xu and Chen, Puhua and Liu, Fang and Li, Yuxuan and Guo, Zhicheng},
  journal={IEEE Transactions on Geoscience and Remote Sensing},
  volume={61},
  pages={1--15},
  year={2023},
  publisher={IEEE}
}

@article{wang2020comprehensive,
  title={A comprehensive survey of optical remote sensing image segmentation methods},
  author={Wang, Yongzhi and Lv, Hua and Deng, Rui and Zhuang, Shengbing},
  journal={Canadian Journal of Remote Sensing},
  volume={46},
  number={5},
  pages={501--531},
  year={2020},
  publisher={Taylor \& Francis}
}

@inproceedings{tang2025aerogen,
  title={AeroGen: Enhancing remote sensing object detection with diffusion-driven data generation},
  author={Tang, Datao and Cao, Xiangyong and Wu, Xuan and Li, Jialin and Yao, Jing and Bai, Xueru and Jiang, Dongsheng and Li, Yin and Meng, Deyu},
  booktitle={2025 IEEE/CVF Conference on Computer Vision and Pattern Recognition (CVPR)},
  pages={3614--3624},
  year={2025},
  organization={IEEE}
}

@article{li2018deep,
  title={Deep learning for remote sensing image classification: A survey},
  author={Li, Ying and Zhang, Haokui and Xue, Xizhe and Jiang, Yenan and Shen, Qiang},
  journal={Wiley Interdisciplinary Reviews: Data Mining and Knowledge Discovery},
  volume={8},
  number={6},
  pages={e1264},
  year={2018},
  publisher={Wiley Online Library}
}

@article{tang2026terragen,
  title={Terragen: A unified multi-task layout generation framework for remote sensing data augmentation},
  author={Tang, Datao and Wang, Hao and Xin, Yudeng and Qiao, Hui and Jiang, Dongsheng and Li, Yin and Yu, Zhiheng and Cao, Xiangyong},
  journal={IEEE Transactions on Geoscience and Remote Sensing},
  year={2026},
  publisher={IEEE}
}

@article{zhang2026uni,
  title={Uni-RS: A Spatially Faithful Unified Understanding and Generation Model for Remote Sensing},
  author={Zhang, Weiyu and Hu, Yuan and Li, Yong and Liu, Yu},
  journal={arXiv preprint arXiv:2601.17673},
  year={2026}
}
\bibliographystyle{iclr2027_conference}

\appendix
\section{Appendix}

\subsection{UniGen Dataset Construction}
\label{app:unigen}


UniGen integrates Git-10M, DIOR-R, and WHU-OPT-SAR to support
text-to-image generation, inpainting, outpainting, controllable
generation, and cross-modal translation. Table~\ref{tab:unigen_composition}
summarizes the source data used for each task. The counts refer to
source images or aligned image pairs. In particular, tasks grouped in
the same row are constructed from the same set of source images,
rather than from separate sets of the stated size.



\vspace{-10pt}

\begin{table}[h]
    \centering
    \caption{Composition of UniGen by source data.}
    \label{tab:unigen_composition}
    \small
    \begin{tabular}{llrrr}
        \toprule
        Generation task
            & Source dataset
            & Images / pairs
            & Train
      
            & Test \\
        \midrule
        Text-to-image
            & Git-10M
            & 2,669,400
            & 2,668,400
     
            & 1,000 \\

        Outpainting
            & Git-10M
            & 2,669,400
            & 2,668,400
       
            & 1,000 \\

        Inpainting
            & Git-10M
            & 199,794
            & 198,795
           
            & 999 \\

        Controllable generation
            & Git-10M, DIOR-R
            & 223,257
            & 210,520
     
            & 12,737 \\

        RGB-to-NIR
            & WHU-OPT-SAR
            & 144,528
            & 144,028
        
            & 500 \\
RGB-to-SAR
            & WHU-OPT-SAR 
            & 127,600
            & 127,100
          
            & 500 \\
        \bottomrule
    \end{tabular}
\end{table}

For text-to-image generation and outpainting, we use a shared subset
of 2,669,400 images from Git-10M. For inpainting and
controllable generation, we use another shared subset of 199,794
Git-10M images. Controllable generation further uses the 23,463 images
in the DIOR-R dataset and their oriented bounding-box
annotations to construct spatial control conditions. For cross-modal translation, we use 144,028 aligned RGB--NIR pairs
and 127,100 aligned RGB--SAR pairs from WHU-OPT-SAR. We convert these
heterogeneous sources into task-specific instructions, visual
conditions, and target images, following the common format
$\{\text{instruction},\text{reference\_image},\text{output\_image}\}$.

\begin{figure*}[h]
  \centering
   \includegraphics[width=1\linewidth]{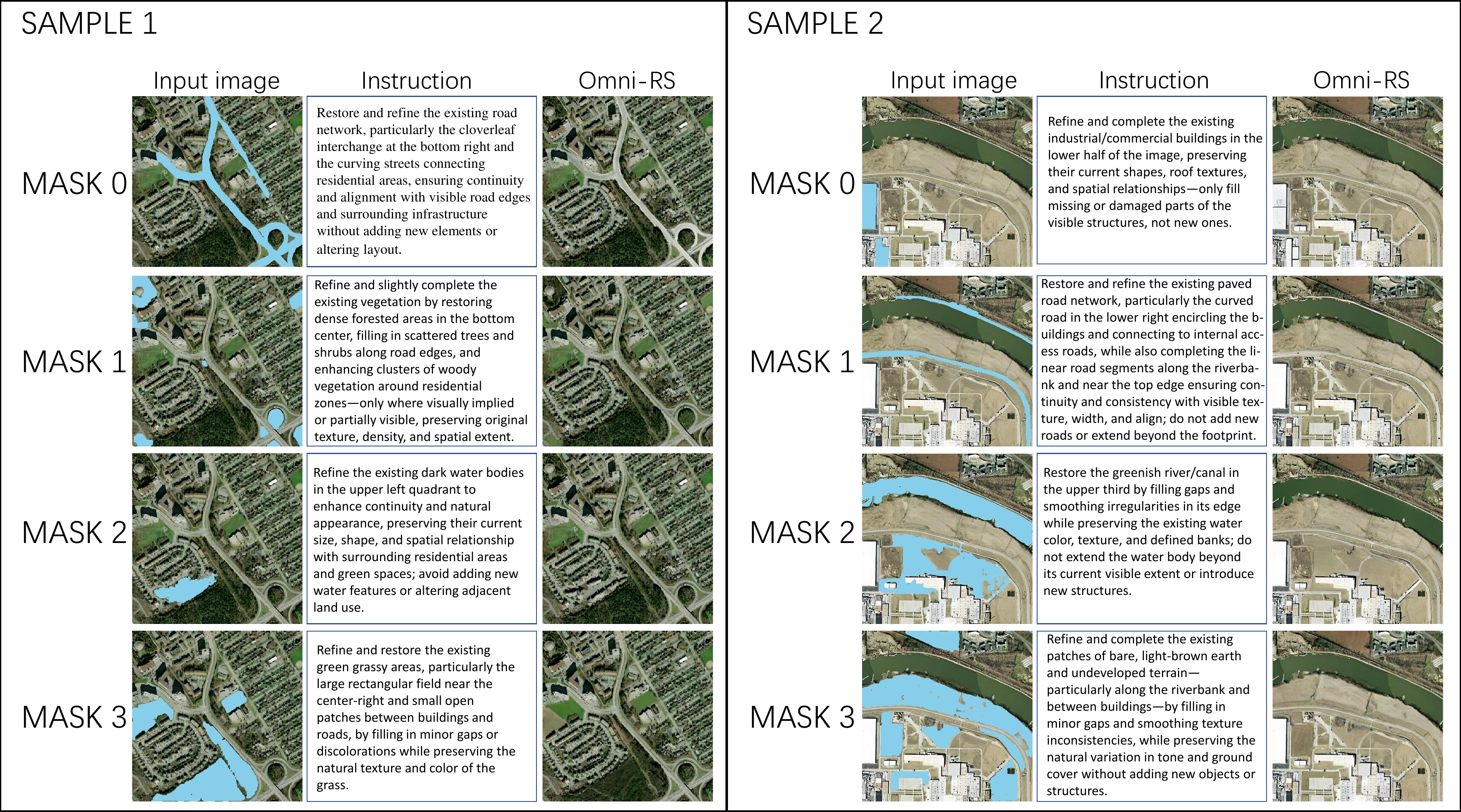}
   \caption{\textbf{Inpainting results of Omni-RS}. For each sample, distinct land-cover categories in the same geographic scene are independently masked and paired with specialized instructions.}
\label{fig:decoupled_masks}
   \label{fig:vis_inpainting}
\end{figure*}

Notably, the inpainting masks in UniGen are generated in a decoupled, category-specific manner, where distinct geographical entities within the same scene are disentangled and preserved as separate masks. As demonstrated in Figure~\ref{fig:vis_inpainting}, Omni-RS accurately restores missing regions under diverse mask contexts and targeted instructions while preserving overarching scene consistency across identical geographic footprints. Crucially, leveraging these disentangled masks, we further construct fine-grained controllable generation pairs via semantic color mapping, achieving synergistic multi-task data construction and synthesis over the exact same geospatial scene.

\begin{figure}[!h]
    \centering
    \includegraphics[
        width=\textwidth,
        height=0.88\textheight,
        keepaspectratio
    ]{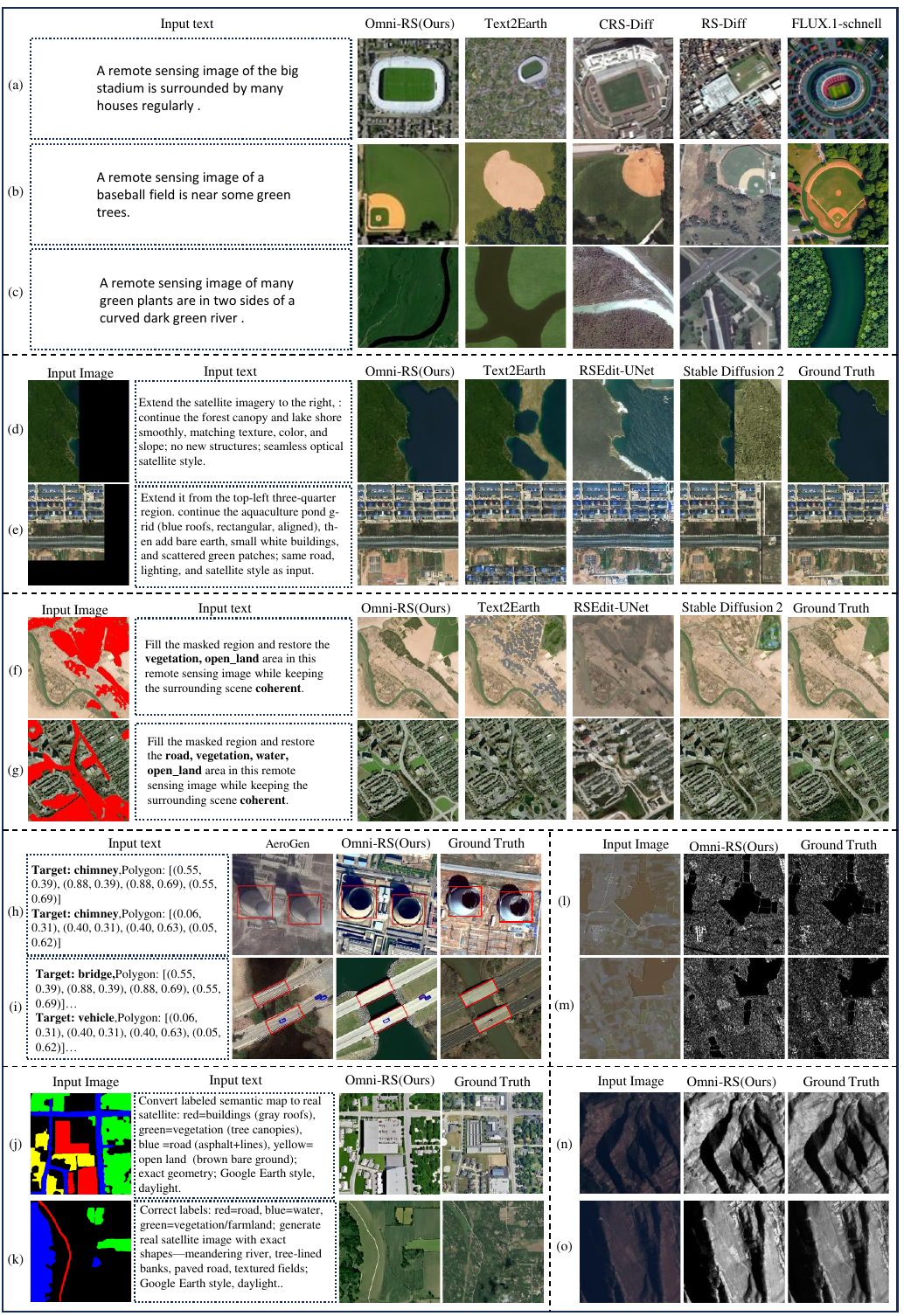}
    \caption{Additional qualitative results of Omni-RS across five
    remote sensing generation tasks. We compare Omni-RS with the
    applicable baseline methods for text-to-image generation,
    outpainting, inpainting, and controllable generation. The input and ground-truth image are provided where applicable.
    Rows (l)--(o) present cross-modal translation results alongside
    their ground-truth targets.}
    \label{fig:additional_generation_results}
\end{figure}

\subsection{Additional Qualitative Comparisons Across Generative Tasks.}

\label{app:qualitative}
As illustrated in Figure~\ref{fig:additional_generation_results}, we present additional qualitative results across all five remote sensing generation tasks, with baseline comparisons where applicable:

\begin{itemize}
    \item \textbf{Text-to-Image Generation (Rows a--c):} Omni-RS accurately captures fine-grained spatial configurations specified in the prompts, such as the regular layout of houses surrounding the stadium and the smooth meandering geometry of the dark green river. In contrast, general-domain baselines like FLUX.1-schnell and specialized models like CRS-Diff often yield distorted spatial structures or non-realistic ground textures.
    \item \textbf{Outpainting and Inpainting (Rows d--g):} Omni-RS exhibits seamless spatial continuity along mask boundaries. In outpainting (Rows d--e), it naturally extends complex geospatial topologies, such as dense forest shorelines and structured aquaculture pond grids, strictly adhering to the prompt description while matching the input style. In inpainting (Rows f--g), thanks to Detailed Visual Conditioning, our model seamlessly reconstructs masked road grids and open terrain without border artifacts or blurriness, noticeably surpassing Stable Diffusion 2 and RSEdit.
    \item \textbf{Controllable Generation (Rows h--k):} For bounding-box-guided synthesis (Rows h--i), Omni-RS strictly follows oriented polygon constraints to place facilities (e.g., chimneys, bridges), outperforming AeroGen in spatial alignment. For semantic color map control (Rows j--k), Omni-RS faithfully adheres to the legend prompts (e.g., distinguishing roads, vegetation, and buildings), synthesizing high-fidelity remote sensing scenes with precise semantic boundaries.
    \item \textbf{Cross-Modal Translation (Rows l--o):} In both RGB$\rightarrow$SAR and RGB$\rightarrow$NIR translations, Omni-RS faithfully preserves macro-scale terrain topographies and ground geometry from the input optical rasters, while realistically rendering sensor-specific physical properties (e.g., radar backscattering patterns and near-infrared reflectance variations), closely matching the ground truth.
\end{itemize}

\begin{figure*}[h]
  \centering
   \includegraphics[width=1\linewidth]{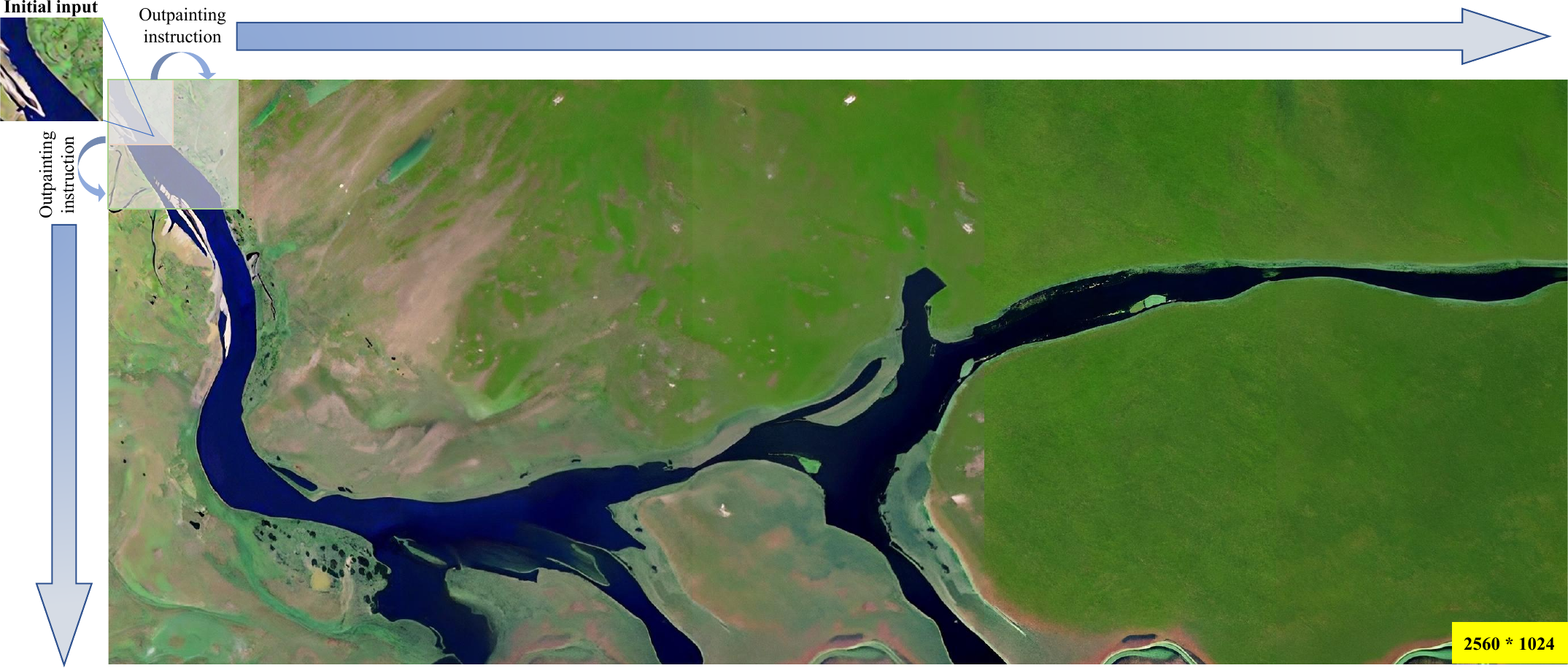}
   \caption{Qualitative results of continuous large-scale outpainting by Omni-RS. Starting from an initial local satellite patch (top-left), the model iteratively expands the geographic canvas along both horizontal and vertical axes guided by outpainting instructions. Omni-RS seamlessly synthesizes an ultra-wide panoramic satellite image ($2560 \times 1024$).}
   \label{fig:vis_outpainting}
\end{figure*}

\subsection{Continuous Outpainting for Large-Scale Geospatial Landscapes.}
\label{app:outpainting}
In practical Earth observation scenarios, downstream applications frequently require extrapolating wide-area, continuous geographical landscapes beyond a single sensor footprint. To evaluate the structural stability and long-range coherence of Omni-RS, we conduct multi-step continuous outpainting starting from a localized input patch ($256 \times 256$), as illustrated in Figure~\ref{fig:vis_outpainting}.

Remarkably, Omni-RS progressively expands the initial crop into an ultra-wide panoramic remote sensing scene with a final canvas size of $2560 \times 1024$. Throughout this iterative generation process, the complex meandering geometry and branch topologies of the central river system remain globally coherent and physically plausible. Furthermore, the surrounding wetland vegetation, sediment textures, and subtle radiometric transitions extend naturally across multiple expansion steps without introducing noticeable stitching boundaries, blurriness, or catastrophic structural collapse. This qualitative demonstration underscores the model's robust spatial generalization and its remarkable ability to preserve long-range contextual continuity in megapixel-scale remote sensing synthesis.

\subsection{Evaluation Metrics}
\label{sec:appendix_metrics}
For understanding, we report VQA accuracy and ROUGE-1 (R-1), ROUGE-L (R-L)~\citep{lin2004rouge}, and METEOR (MR)~\citep{banerjee2005meteor} for region understanding. Text-to-image generation is evaluated using Fr{\'e}chet Inception Distance (FID)~\citep{heusel2017gans} and CLIP similarity~\citep{radford2021learning}. Inpainting and outpainting are evaluated using FID, Kernel Inception Distance (KID)~\citep{binkowski2018demystifying}, and Learned Perceptual Image Patch Similarity (LPIPS)~\citep{zhang2018unreasonable}. For controllable generation, we report YOLO Score~\citep{li2021image} and Classification Accuracy Score (CAS)~\citep{ravuri2019classification}. Cross-modal translation is evaluated using MAE, Structural Similarity Index (SSIM)~\citep{wang2004image}, and Peak Signal-to-Noise Ratio (PSNR) for RGB-to-NIR, and MAE, RMSE, and PSNR for RGB-to-SAR.


\end{document}